# ReRe: A Lightweight Real-time Ready-to-Go Anomaly Detection Approach for Time Series

Ming-Chang Lee[1], Jia-Chun Lin[2], and Ernst Gunnar Gran[3]

[1,2,3]Department of Information Security and Communication Technology, Norwegian University of Science and Technology, Ametyst-bygget, 2815 Gjøvik, Norway
[3]Simula Research Laboratory, 1364 Fornebu, Norway

[1] ming-chang.lee@ntnu.no
[2]jia-chun.lin@ntnu.no
[3]ernst.g.gran@ntnu.no

4th December 2022

This work was supported by the project eX[3] - *Experimental Infrastructure for Exploration of Exascale Computing* funded by the Research Council of Norway under contract 270053 and the scholarship under project number 80430060 supported by Norwegian University of Science and Technology.

# ReRe: A Lightweight Real-time Ready-to-Go Anomaly Detection Approach for Time Series

Ming-Chang Lee
*Department of Information Security and Communication Technology*
*Norwegian University of Science and Technology*
Gjøvik, Noway
ming-chang.lee@ntnu.no

Jia-Chun Lin
*Department of Information Security and Communication Technology*
*Norwegian University of Science and Technology*
Gjøvik, Noway
jia-chun.lin@ntnu.no

Ernst Gunnar Gran
*Department of Information Security and Communication Technology*
*Norwegian University of Science and Technology,* Gjøvik, Noway
*Simula Research Laboratory,* Fornebu, Norway
ernst.g.gran@ntnu.no

***Abstract*—Anomaly detection is an active research topic in many different fields such as intrusion detection, network monitoring, system health monitoring, IoT healthcare, etc. However, many existing anomaly detection approaches require either human intervention or domain knowledge, and may suffer from high computation complexity, consequently hindering their applicability in real-world scenarios. Therefore, a lightweight and ready-to-go approach that is able to detect anomalies in real-time is highly sought-after. Such an approach could be easily and immediately applied to perform time series anomaly detection on any commodity machine. The approach could provide timely anomaly alerts and by that enable appropriate countermeasures to be undertaken as early as possible. With these goals in mind, this paper introduces ReRe, which is a Real-time Ready-to-go proactive Anomaly Detection algorithm for streaming time series. ReRe employs two lightweight Long Short-Term Memory (LSTM) models to predict and jointly determine whether or not an upcoming data point is anomalous based on short-term historical data points and two long-term self-adaptive thresholds. Experiments based on real-world time-series datasets demonstrate the good performance of ReRe in real-time anomaly detection without requiring human intervention or domain knowledge.**

## I. INTRODUCTION

A time series is a series of data points generated by for instance system monitoring tools, smart meters, stock exchange, or IoT devices, etc., where the data points are evenly indexed in time order. Anomaly detection refers to the identification of abnormal or novel events embedded in a time series [1]. An abnormal event is usually called an anomaly, which could be a point anomaly or a temporal anomaly. A point anomaly, also called an outlier, happens when a single data point is outside the expected range, whereas a temporal anomaly occurs when the pattern of the time series changes and lasts for a while [2].

In the last decade, a number of approaches and methods have been introduced for time series anomaly detection in different areas such as intrusion detection, IP networks, software bug detection, system health detection, smart homes, healthcare, etc. Examples include [2][5][9-14]. However, most approaches require either domain knowledge or human intervention. For instances, they might require humans to collect sufficient and representative training data, manually label training data, conduct an off-line training process, understand the distribution of the target time series, perform parameter tuning, and/or determine appropriate thresholds empirically. These requirements consequently limit the applicability and usefulness of these approaches in practice.

It would be highly valuable to facilitate a ready-to-go and self-adaptive anomaly detection approach for time series where human intervention and domain knowledge are not needed. Such an approach could be easily and immediately utilized and could adapt to pattern and distribution changes in any time series. Furthermore, it might also be highly valuable to provide a real-time and proactive anomaly detection approach for time series since such an approach helps to trigger prompt troubleshooting and enables appropriate countermeasures to be undertaken as soon as possible. Last but not least, it would be highly beneficial to offer a lightweight anomaly detection approach with all above-mentioned features because such an approach can be deployed on any commodity machine, such as desktops, laptops, and mobile phones.

In this paper, we introduce ReRe, which stands for Real-time Ready-to-go Proactive Time Series Anomaly Detection based on Long Short-Term Memory (LSTM) [7]. The goal of ReRe is to provide accurate, real-time, and lightweight anomaly detection for time series without requiring domain knowledge or human intervention. In other words, ReRe must work completely by itself to learn the data distribution of the target time series in an online manner, dynamically determine its detection threshold to adapt to pattern changes in the target time series, and detect any anomalous data points either proactively or on time. To achieve these goals, ReRe employs two LSTM models and two long-term self-adaptive detection thresholds to predict and jointly decide if the upcoming data point in the target time series is anomalous or not. The two LSTM models run in parallel with a simple network structure and are always trained with short-term historical data points. These features enable lightweight and real-time LSTM training and anomaly detection. By dynamically and individually adjusting each threshold over time, the two LSTM models are able to tolerate minor pattern changes and detect anomalies in the target time series. Note that the two thresholds are determined differently. One is determined by taking all data points collected so far into consideration. The other one is determined by taking all data points that are considered normal into consideration. The purpose is to provide two levels of sensitivity for accurate anomaly detection.

To demonstrate the detection performance of ReRe, we conducted experiments based on real-world time series datasets provided by the Numenta Anomaly Benchmark (NAB) [3] and the Yahoo benchmark datasets [15]. We compared ReRe with three state-of-the-art anomaly detection approaches. The results show that ReRe outperforms the other approaches in precision, recall, and F-score. In addition, the results also demonstrate that ReRe is lightweight, computationally efficient, and able to conduct anomaly detection in real-time. The contributions of this paper are as follows:

1. The proposed ReRe is a generic, ready-to-go, and completely unsupervised learning approach. It can be easily and immediately applied to detect anomalies in any streaming time series without knowing the distribution of the target time series or requiring human effort to pre-train a learning model, pre-build a data model, tune parameters, or set a detection threshold manually.
2. ReRe is both lightweight and cost-effective due to the employed simple LSTM network structure and the short-term Look-Back and Predict-Forward strategy [16]. These features enable ReRe to provide anomaly detection in real-time.
3. ReRe is able to tolerate and adapt to pattern changes in the target time series due to its LSTM retraining characteristic and the two long-term self-adaptive detection thresholds.

The rest of the paper is organized as follows: Section II describes related work. In Section III, we introduce the details of ReRe. Section IV presents and discusses the experiments and the corresponding results. In Section V, we conclude this paper and outline future work.

## II. Related Work

Over the years, a number of anomaly detection approaches have been introduced. Statistical learning approaches are one of the categories. These approaches work by fitting a statistical model to a given set of normal data and then use the model to determine whether an upcoming data point fits this model or not. If the data point has a low probability to be generated from the model, it is considered anomalous. For instance, Twitter has proposed two anomaly detection algorithms, called AnomalyDetectionTs (ADT for short), and AnomalyDetectionVec (ADV for short). Both of them have been implemented and included in an open-source R package [4]. ADT is designed to detect one or more statistically significant anomalies in a given time series, while ADV is designed to detect one or more statistically significant anomalies in a given vector of observations without timestamp information. Since ADT and ADV are statistical based, they need sufficient amount of data points in the target time series and consequently might not be an appropriate solution to detect anomalies in streaming time series. In addition, these two approaches are parameter sensitive. They require human experts to set appropriate values to their parameters in order to achieve good detection performance.

Luminol [12] is another anomaly detection approach proposed by LinkedIn for time series. Luminol is implemented as an open-source Python library for identifying anomalies in real user monitoring (RUM) data for LinkedIn pages and applications. Given a time series, Luminol calculates an anomaly score for each data point in the time series. If a data points has a high score, it means that this data point is likely to be anomalous as compared with other data points in the time series. In other words, human experts still need to further determine which data points are anomalies based on their experiences. In addition, Luminol suffers from similar issues as ADT and ADV since it is also statistical based.

Machine learning approaches represent another category of anomaly detection. Most approaches belonging to this category require either domain knowledge or human intervention. For example, Yahoo introduced EGADS [9] to detect anomalies on time series based on a collection of anomaly detection and forecasting models. However, EGADS requires to model the target time series so as to predict a data value later used by its anomaly detection module and its altering module. Lavin and Ahmad [2] proposed Hierarchical Temporal Memory (HTM) to capture changing patterns in time series. However, HTM requires 15% of a training dataset to be non-anomalous so that it can used this data to train its neural network. Different from EGADS and HTM, the approach proposed in this paper (i.e., ReRe) does not have these requirements.

Siffer et al. [10] proposed a time series anomaly detection approach based on Extreme Value Theory. This approach makes no assumption on the distribution of time series and requires no threshold manually set by humans. However, this approach needs a long time period to do necessary calibration before conducting anomaly detection. According to [10], the calibration process needs at least 1000 data points, which is much longer than the probation period required by ReRe. Greenhouse [6] is a zero-positive anomaly detection algorithm for time series based on LSTM. Greenhouse requires all data points in its training datasets to be non-anomalous, making Greenhouse a kind of supervised learning approach. During the training phase, Greenhouse adopts a Look-Back and Predict-Forward strategy to detect anomalies. For a given time point $t$, a window of most recently observed values of length $B$ is used as "Look-Back" to predict a subsequent window of values of length $F$ as "Predict-Forward". This feature enables Greenhouse to adapt to pattern changes in the training data. However, if the training data is not representative, Greenhouse might not be able to capture and accommodate pattern changes in real-world time series.

RePAD [16] is a real-time time series anomaly detection approach also based on the Look-Back and Predict-Forward strategy. RePAD utilizes a single LSTM model trained with short-term historic data points to be the predictor and detector. Together with the LSTM model, a dynamically adjusted long-term detection threshold is utilized to determine if each data point in the target time series is anomalous or not. According to the experiment results shown in [16], RePAD is able to detect anomalies either proactively or on time, but RePAD suffers from some undesirable false positives. Different from RePAD, ReRe employs two LSTM models and two long-term self-adaptive thresholds to detect anomalies in a parallel manner. The two thresholds provide two levels of detection sensitivity aiming to keep a high true positive rate and a low false positive rate. The details of ReRe will be introduced in the next section, and the comparison between ReRe, ADT, ADV, and RePAD will be shown in Section IV.

## III. The Details of ReRe

As stated earlier, ReRe utilizes the Look-Back and Predict-forward strategy based on short-term historic data points. More specifically, ReRe utilizes two LSTMs to individually predict each data point in the target time series based on the data values observed at the past $b$ continuous time points, and then determine if the next data point is anomalous or not. Note that $b$ is called the Look-Back parameter, and that $b$ is a small integer, implying that training data used to train the two LSTMs is small in size (i.e., $b$ data points). Therefore, a simple network structure should be sufficient for the two LSTMs. Due to this, each LSTM consists of only one hidden layer with 10 hidden units. Each LSTM is always trained with the learning rate of 0.15, which enables a fast learning speed and provides a satisfactory learning result. With respect to epoch (which is defined as one forward pass and one backward pass of all the training data), it is clear that too many epochs might overfit the training data, whereas too few epochs may underfit the training data. To address this issue, ReRe employs Early Stopping [8] to automatically determine the number of epochs for each LSTM. In this paper, Early Stopping always chooses a number between 1 and 50.

ReRe consists of one main function and two sub-procedures. One is called Detector 1, and the other is called Detector 2. Let $t$ be the current time point. Note that $t$ starts at 0, which is the moment when ReRe is launched. As illustrated in Fig. 1, when ReRe is launched, this approach will go through a short probation period with a length of $2b+1$ time points (see lines 3 to 11, Fig. 1). During this period, ReRe keeps training an LSTM model based on the past $b$ observed data points, uses the corresponding LSTM model to predict the value of the data point at the next time point, and then derives the corresponding AARE values. Note that AARE stands for Average Absolute Relative Error. A low AARE value indicates that the predicted values are close to the observed values. During the probation period, everything created or generated will be duplicated and used later by Detector 1 and Detector 2.

Whenever time advances and $t \geq 2b+1$, ReRe invokes Detector 1 and Detector 2 to separately perform anomaly detection by passing $t$ and $v_t$ to both of them. If both Detector 1 and Detector 2 return that $v_t$ is abnormal (See line 14 of Fig. 1), ReRe concludes that $v_t$ is an anomaly and immediately reports it to trigger troubleshooting or countermeasures.

Fig. 2 shows the algorithm of Detector 1 where $M_1$ is a duplicate of the LSTM model created in the probation period. Whenever receiving $t$ and $v_t$ from ReRe, Detector 1 calculates $AARE'_t$ based on Equation 1:

$$AARE'_t = \frac{1}{b} \cdot \sum_{y=t-b+1}^{t} \frac{\left| v_y - \widehat{v'_y} \right|}{v_y} \text{ , } t \geq 2b-1 \qquad (1)$$

where $v_y$ is the observed data value at time point $y$, and $\widehat{v'_y}$ is the forecast data value predicted by $M_1$ at $y$, where $y = t-b+1, t-b+2, \ldots, t$. After that, as shown by lines 3 to 9 of Fig. 2, Detector 1 calculates a detection threshold, denoted by $thd_1$, by considering all previously calculated AARE values (i.e., $AARE'_b$, $AARE'_{b+1}$, …, $AARE'_t$) based on the Three-Sigma Rule [5], which is a commonly used rule for anomaly detection.

If $AARE'_t$ is smaller than or equal to $thd_1$ (see line 10 of Fig. 2), it means that $v_t$ is similar to previous data points. In this case, Detector 1 replies to ReRe that $v_t$ is normal and keeps using the current LSTM model (i.e., $M_1$) to predict the next data point $\widehat{v'_{t+1}}$. However, if $AARE'_t$ is greater than $thd_1$, implying that either the data pattern of the target time series has changed or an anomaly might happen, Detector 1 retrains its LSTM model by taking the most recent $b$ data points, i.e., $[v_{t-b}, v_{t-b+1} \ldots, v_{t-1}]$, as the training data. After that, Detector 1 uses this new LSTM model to re-predict $\widehat{v'_t}$ and then re-calculates the corresponding $AARE'_t$ (see lines 13 to 15 of Fig. 2).

**ReRe algorithm**
**Input**: Data points in the target time series
**Output**: Anomaly notifications

1: **While** time has advanced {
2: Let $t$ be the current time point and $v_t$ be the data point collected at $t$;
3: **if** $t \geq b-1$ & $t < 2b-1$ { // i.e., $2 \leq t < 5$, if $b = 3$
4: Train an LSTM model called $M_1$ with training data $[v_{t-b+1}, v_{t-b+2} \ldots, v_t]$;
5: Use $M_1$ to predict $\widehat{v'_{t+1}}$;
6: Let $M_2$ and $\widehat{v''_{t+1}}$ be the duplicates of $M_1$ and $\widehat{v'_{t+1}}$ respectively;}
7: **else if** $t \geq 2b-1$ & $t < 2b+1$ { //i.e., $5 \leq t < 7$, if $b = 3$
8: Calculate $AARE'_t$ based on Equation 1;
9: Train $M_1$ with training data $[v_{t-b+1}, v_{t-b+2} \ldots, v_t]$;
10: Use $M_1$ to predict $\widehat{v'_{t+1}}$;
11: Let $AARE''_t$, $M_2$, and $\widehat{v''_{t+1}}$ be the duplicates of $AARE'_t$, $M_1$, and $\widehat{v'_{t+1}}$ respectively;}
12: **else if** $t \geq 2b+1${ //The probation period has passed.
13: Invoke Detector 1 and Detector 2 by passing $t$ and $v_t$ to both of them;
14: **if** both Detector 1 and Detector 2 consider $v_t$ abnormal{
15: $v_t$ is reported as an anomaly immediately;}}}

Fig. 1. The algorithm of ReRe.

**Detector 1**
**Input**: $t$ and $v_t$
**Output**: A returned message to ReRe

1: Obtain $t$ and $v_t$ from ReRe;
2: Calculate $AARE'_t$ based on Equation 1;
3: Let $sum = 0$, $x = b - 1$, and $c = 0$;
4: **while** $x \leq t - b${ $sum = sum + AARE'_{b+x}$; $x = x + 1$;}
5: $\mu_1 = sum/(t - b + 1)$;
6: Reset $sum$ and $x$ to be zero;
7: **while** $x \leq t - b${ $sum = sum + (AARE'_{b+x} - \mu_1)^2$; $x = x + 1$;}
8: $\sigma_1 = \sqrt{sum/(t - b + 1)}$;
9: $thd_1 = \mu_1 + 3\sigma_1$;
10: **if** $AARE'_t \leq thd_1${
11: Return message "$v_t$ is normal" to ReRe; Use $M_1$ to predict $\widehat{v'_{t+1}}$ ;}
12: **else**{
13: Retrain a new LSTM by taking $[v_{t-b}, v_{t-b+1} \dots, v_{t-1}]$ as the training data;
14: Use the new trained LSTM to re-predict $\widehat{v'_t}$;
15: Re-calculate $AARE'_t$ using Equation 1;}
16: **if** $AARE'_t \leq thd_1${
17: Return message "$v_t$ is normal" to ReRe; Replace $M_1$ with the new trained LSTM;}
18: **else**{
19: Return message "$v_t$ is abnormal" to ReRe;}}

Fig. 2. The algorithm of Detector 1.

**Detector 2**
**Input**: $t$ and $v_t$
**Output**: A returned message to ReRe

1: Obtain $t$ and $v_t$ from ReRe;
2: Calculate $AARE''_t$ based on Equation 2;
3: Let $sum = 0$, $x = b - 1$ and $c = 0$;
4: **while** $x \leq t - b${
5: **if** $v_{b+x}$ was considered normal by Detector2{ $sum = sum + AARE''_{b+x}$; $c = c + 1$;}
6: $x = x + 1$;}
7: $\mu_2 = sum/c$;
8: Reset $sum$ and $x$ to be zero;
9: **while** $x \leq t - b${
10: **if** $v_{b+x}$ was considered normal by Detector2{ $sum = sum + (AARE''_{b+x} - \mu_2)^2$; $c = c + 1$;}
11: $x = x + 1$;}
12: $\sigma_2 = \sqrt{sum/c}$;
13: $thd_2 = \mu_2 + 3\sigma_2$;
14: **if** $AARE''_t \leq thd_2${
15: Return message "$v_t$ is normal" to ReRe; Use $M_2$ to predict $\widehat{v''_{t+1}}$;}
16: **else**{
17: Retrain an LSTM by taking $[v_{t-b}, v_{t-b+1} \dots, v_{t-1}]$ as the training data;
18: Use the new trained LSTM model to predict $\widehat{v''_t}$;
19: Re-calculate $AARE''_t$ using Equation 2;
20: **if** $AARE''_t \leq thd_2${
21: Return message "$v_t$ is normal" to ReRe; Replace $M_2$ with the new trained LSTM model;}
22: **else**{
23: Return message "$v_t$ is abnormal" to ReRe;}}

Fig. 3. The algorithm of Detector 2.

If the new $AARE'_t$ is smaller than or equal to $thd_1$ (see line 16), Detector 1 considers that the data pattern in the time series has slightly changed and that $v_t$ is normal. In this case, Detector 1 replaces $M_1$ with this new trained LSTM model to adapt to the pattern change. On the contrary, if the new $AARE'_t$ is still larger than $thd_1$ (see line 18 of Fig. 2), Detector 1 considers that $v_t$ is abnormal since the LSTM trained with the most recent data points is still unable to accurately predict $v_t$. At this time point, a warming message is immediately sent to ReRe for further evaluation.

Fig. 3 illustrates the algorithm of Detector 2, which is similar to that of Detector 1, except when it comes to how $AARE''_t$ and detection threshold $thd_2$ are calculated. Note that Detector 2 calculates $AARE''_t$ based on Equation 2:

$$AARE''_t = \frac{1}{b} \cdot \sum_{y=t-b+1}^{t} \frac{\left|v_y - \widehat{v''_y}\right|}{v_y}, t \geq 2b - 1 \quad (2)$$

where $\widehat{v''_y}$ is the data value predicted by Detector 2 for time point $y$. As shown from lines 3 to 13 of Fig. 3, if a data point $v_{b+x}$ (where $x = 0,1, \dots, t-b$) was considered normal by Detector 2, the corresponding AARE value (i.e., $AARE''_{b+x}$) will be included to calculate $thd_2$. In other words, all ARRE values associated with the data points that are considered abnormal by Detector 2 will be excluded from the calculation of $thd_2$. This is the key reason why $thd_2$ is different from $thd_1$, and why $M_2$ behaves differently from $M_1$.

Note that Detector 1 and Detector 2 do not need to retrain their LSTM models for every new point in time. As long as their current LSTM models are able to make a prediction such that the corresponding AARE values are under the corresponding detection thresholds, the LSTM models can be used again. This feature enables ReRe to remain lightweight and provide anomaly detection in real-time.

## IV. Experiment Results

In this section, we evaluate the detection performance of ReRe by designing three experiments where we compare ReRe with RePAD [16], ADT [4] and ADV [4]. Recall that ADT and ADV are two open-source statistical-based approaches introduced by Twitter, while RePAD is a real-time time series anomaly detection approach. Recall that Luminol [12] is also an open-source anomaly detection approach. However, Luminol only produces a score for each data point in the target time series without being able to indicate which one is anomaly. Due to this reason, luminol was not chosen for the comparison. All the experiments were performed on a commodity laptop running macOS 10.15.1 with 2.6 GHz 6-Core Intel Core i7 and 16GB DDR4 SDRAM.

TABLE I. Three real-world time-series datasets

| Name | Time Period | # of data points | Time Interval |
|---|---|---|---|
| CPU-cc2 | From 2014-04-10, 00:04 to 2014-04-24, 00:09 | 4032 | 5 min |
| CPU-c53 | From 2014-02-14, 14:30 to 2014-02-28, 14:30 | 4032 | 5 min |
| real47 | Not provided | 1427 | 1 hour |

In the first two experiments, two real-world time series datasets called ec2-cpu-utilization-825cc2 and rds-cpu-utilization-cc0c53 were chosen from NAB [2] to evaluate the four abovementioned approaches. These two datasets are separately abbreviated as CPU-cc2 and CPU-c53 in this paper. The last dataset, called real47, was selected from the Yahoo benchmark datasets [15]. TABLE I lists the details of these datasets. Note that the interval time between data points in the first two datasets are 5 minutes, whereas the interval time between data points in the last dataset is one hour.

TABLE II shows all parameter settings for each of the four approaches. Note that the Look-Back parameter (i.e., $b$) is the only parameter we need to set for both ReRe and RePAD. Since both approaches are based on short-term historic data points to detect anomalies, we followed the setting used in [16] (i.e., $b = 3$) for both ReRe and RePAD in all the experiments. On the other hand, it is a requirement to set parameter $k$ (which indicates the maximum number of anomalies to detect) for both ADT and ADV. We followed the setting mentioned in [4] and set $k$ as 0.02. Furthermore, it is also a requirement to set the *period* parameter for ADV. In the first two experiments, we set this parameter as 1440 according to the setting suggested by [4]. However, in the last experiment, the same setting was unaccepted by ADV since the value of 1440 is more than the total length of the real47 dataset. In order to fairly evaluate ADV, we set this parameter as 300, 500, and 700 to see how ADV performs under each of these settings. It is worth noting that ADT was unable to execute the last experiment since ADT requires the target dataset to be two times longer than real47.

TABLE II. Parameter Setting of the Three Experiments

| Approach | Experiment 1 | Experiment 2 | Experiment 3 |
|---|---|---|---|
| ReRe | $b = 3$ | $b = 3$ | $b = 3$ |
| RePAD | $b = 3$ | $b = 3$ | $b = 3$ |
| ADT | $k = 0.02$ | $k = 0.02$ | unexecutable |
| ADV | $k = 0.02$ *period*=1440 | $k = 0.02$ *period*=1440 | $k = 0.02$ *period*=300, 500, 700 |

### A. *Experiment 1*

Fig. 4 illustrates all the data points in the CPU-cc2 dataset and the detection results of the four approaches on the dataset. In this dataset, there are two anomalies labeled by human experts, and both of them are marked as red circles in Figs. 4 and 5.

When ReRe and RePAD were individually employed, they made some false detections in the beginning. However, this situation did not happen frequently after ReRe and RePAD learned the data pattern in the time series. We can see that ReRe then produced less false detections than RePAD, especially in the period before the first real anomaly, implying that employing the two LSTM models with the two long-term self-adaptive detection thresholds is able to mitigate false positives.

In order to clearly view the detection results of all the approaches, Fig. 5 depicts a close-up of the detection results. Apparently, both ReRe and RePAD are the only two approaches that are able to detect the first anomaly on time (i.e., when this anomaly occurs) and to detect the second anomaly proactively around 5 minutes earlier than the occurrence of the second anomaly. On the contrary, both ADT and ADV are only able to detect the second anomaly on time, without being able to detect the first anomaly at all. Besides, ADT and ADV made a lot of false positives after the occurrence of the second anomaly, implying that these two approaches are unable to adapt to the pattern change in the time series.

TABLE III. The detection performance of the four approaches on the CPU-cc2 dataset.

| Approach | Precision | | Recall | | F-score | |
|---|---|---|---|---|---|---|
| | *K*=0 | *K*=7 | *K*=0 | *K*=7 | *K*=0 | *K*=7 |
| ReRe | 0.0513 | 0.5263 | 1 | 1 | 0.0976 | 0.6896 |
| RePAD | 0.0487 | 0.5000 | 1 | 1 | 0.0929 | 0.6667 |
| ADT | 0.0125 | 0.1648 | 0.5 | 0.5 | 0.0244 | 0.2479 |
| ADV | 0.0125 | 0.1648 | 0.5 | 0.5 | 0.0244 | 0.2479 |

TABLE III summaries the precision, recall, and F-score of all the approaches on the CPU-cc2 dataset. Recall that F-score is defined as the weighted harmonic mean of the precision and recall of the test as below:

$$\text{F-score} = 2 \times \frac{\text{precision}\times\text{recall}}{\text{precision}+\text{recall}} \tag{3}$$

The F-score reaches the best value, meaning perfect precision and recall, at a value of 1. The worst F-score would be a value of 0, implying the lowest precision and the lowest recall. Note that both ReRe and RePAD are capable of proactive anomaly detection, thus adopting traditional point-wise metrics to measure them is unsuitable and unfair. Therefore, we adopt and revise the evaluation method proposed by [14] to provide appropriate and fair comparison. More specifically, if any anomaly occurring at time point $t$ can be detected within a time period ranging from time point $t - K$ to time point $t + K$, we say that this anomaly is correctly detected. From TABLE III, it is clear that when the traditional point-wise metrics (i.e., Precision at $K = 0$, Recall at $K = 0$, and F-score at $K = 0$) are used, ReRe performs the best among the four approaches. However, these metrics cannot reflect its capability when it comes to proactive detection. When we followed [14] and set $K$ as 7, we can see the precision, recall, and F-score of each approach all increase. Nevertheless, ReRe still outperforms the rest of the approaches.

TABLES IV and V list the time performance of both ReRe and RePAD on detecting anomalies in the CPU-cc2 dataset. Note that ADT and ADV were not included in this comparison since both of them are statistical based without using LSTM. Apparently, ReRe needs to retrain its LSTM models more frequently than RePAD due to the employment of two LSTMs. However, if we take the whole dataset into consideration, the LSTM retraining ratio of ReRe is only 8.84% (=356/4028), which is very low. Due to this reason, the average time required by ReRe to detect each data point (i.e., 0.039 sec) is slightly higher than that required by RePAD (i.e., 0.026 sec). Note that the detection time for each data point includes both the corresponding LSTM retraining time (if the retraining is necessary) and the corresponding detection time. The results confirm that employing the two LSTMs does not introduce significant computational complexity and load to the underlying laptop, indicating that ReRe is lightweight and able to provide anomaly detection in real-time.

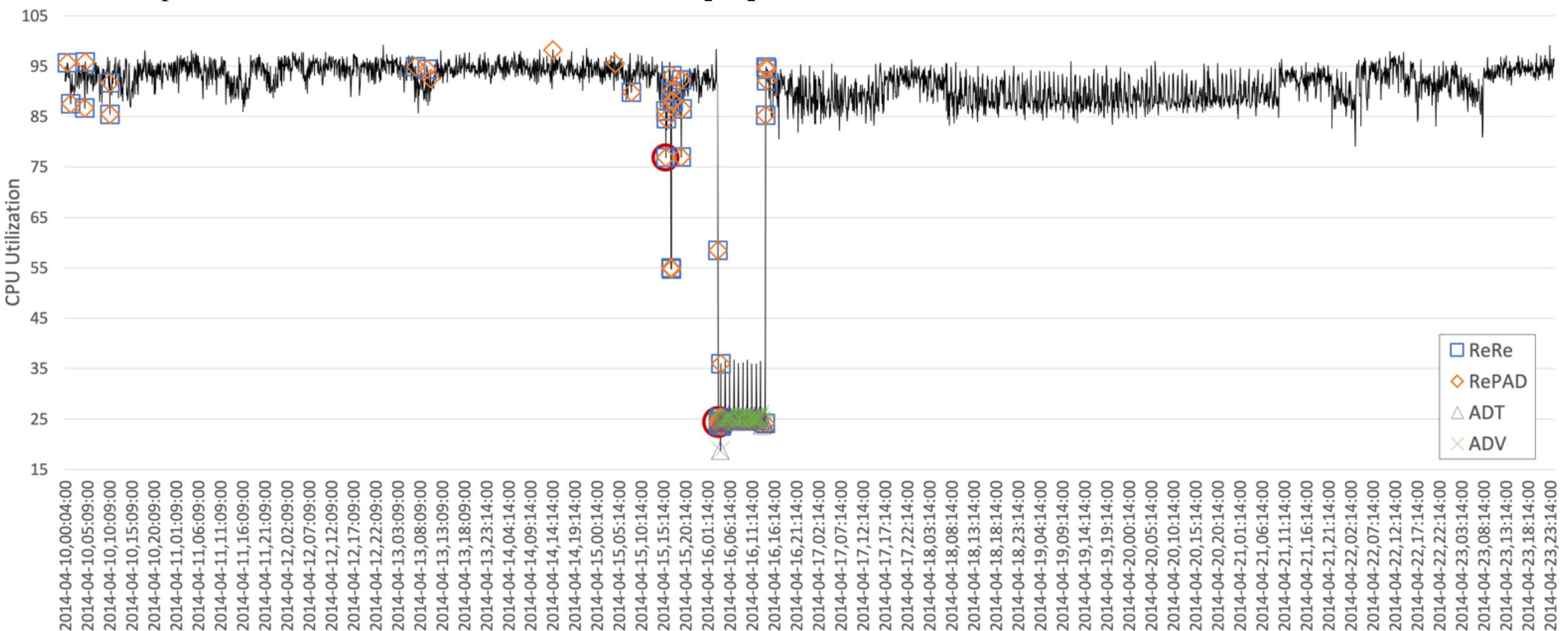

Fig. 4. The detection results of ReRe, RePAD, ADT, and ADV on the CPU-cc2 dataset. Note that this dataset has two anomalies labeled by human experts, marked as red circles.

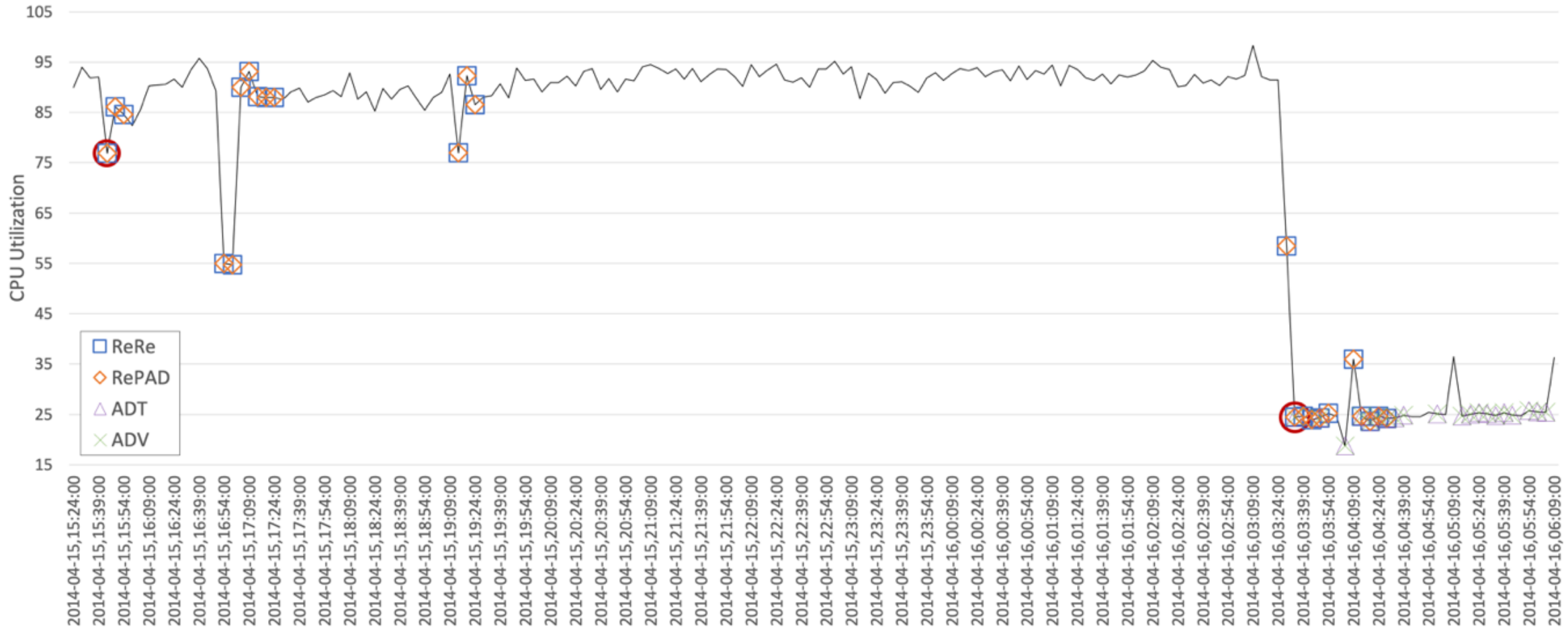

Fig. 5. A close-up of the detection results for the first and second anomalies on the CPU-cc2 dataset.

TABLE IV. THE LSTM RETRAINING PERFORMANCE OF RERE AND REPAD ON THE CPU-CC2 DATASET.

| Approach | # of data points that requires LSTM Retraining | LSTM retraining ratio |
|---|---|---|
| ReRe | 356 | 8.84% (=356/4028) |
| RePAD | 83 | 2% (=83/4028) |

TABLE V. THE TIME CONSUMPTION OF RERE AND REPAD ON THE CPU-CC2 DATASET.

| Approach | Average Detection Time (sec) | Standard Deviation (sec) |
|---|---|---|
| ReRe | 0.039 | 0.075 |
| RePAD | 0.026 | 0.050 |

### B. *Experiment 2*

In the second experiment, the CPU-c53 dataset was used to evaluate the four approaches. This dataset contains two real anomalies labeled by human experts. Fig. 6 illustrates the whole dataset, the two anomalies (marked as red circles), and all detection results of all the four approaches. In order to clearly view the detection results of all the approaches, Fig. 7 depicts a close-up of the detection results for the first and the second anomalies on CPU-c53. Similar to the first experiment, both ReRe and RePAD generated some false positives before they learned the data pattern of the time series. This is unavoidable since ReRe and RePAD learn the time series entirely by themselves without any domain knowledge or human intervention. TABLE VI summaries the detection performance of each approach under two different values of $K$. When $K$ is set to 0, both ADT and ADV have the highest value of recall (i.e., 1) since they are able to detect the two anomalies on time. However, they produce a high number of false positives (which can be seen from both Figs. 6 and 7), which considerably impact their performance in precision and F-score. These false positives also demonstrate that both ADT and ADV are unable to adapt to the pattern changes in the time series.

Apparently, ReRe and RePAD perform better than ADT and ADV in terms of precision and F-score when $K$=0, which as mentioned earlier is not a suitable and fair evaluation for ReRe and RePAD. When $K$ is enlarged to 7, which is a suggested measure according to [14], we can see that ReRe outperforms the other three approaches in all the metrics due to its good performance when it comes to true positives, false positives, and false negatives. Overall speaking, ReRe offers higher precision, recall, and F-score than the other three approaches, no matter if $K$ is set to 0 or 7. The detection performance of ReRe is satisfactory, given that ReRe learns and adapts to the data patterns of the time series completely by itself without obtaining knowledge from the dataset or help from human in advance.

TABLE VI. THE DETECTION PERFORMANCE OF THE FOUR APPROACHES ON THE CPU-C53 DATASET.

| Approach | Precision | | Recall | | F-score | |
|---|---|---|---|---|---|---|
| | $K$=0 | $K$=7 | $K$=0 | $K$=7 | $K$=0 | $K$=7 |
| ReRe | 0.045 | 0.533 | 0.5 | 1 | 0.0825 | 0.695 |
| RePAD | 0.037 | 0.457 | 0.5 | 1 | 0.0689 | 0.627 |
| ADT | 0.025 | 0.174 | 1 | 1 | 0.0487 | 0.296 |
| ADV | 0.025 | 0.174 | 1 | 1 | 0.0487 | 0.296 |

Table VII and VIII list the time performance of ReRe and RePAD on detecting anomalies in the CPU-c53 dataset. When ReRe is employed, it requires to retrain its LSTM models at 111 data points, which is approximately 1.88 (=111/59) times of that required by RePAD. Nevertheless, the LSTM retraining ratio of ReRe is very low since it is only 2.76% (=111/4028), implying that the overhead introduced by ReRe is insignificant. This also explains why the average detection time taken by ReRe is just a little longer than the one taken by RePAD. In other words, the results confirm that ReRe is a cost-effective and time-efficient anomaly detection approach.

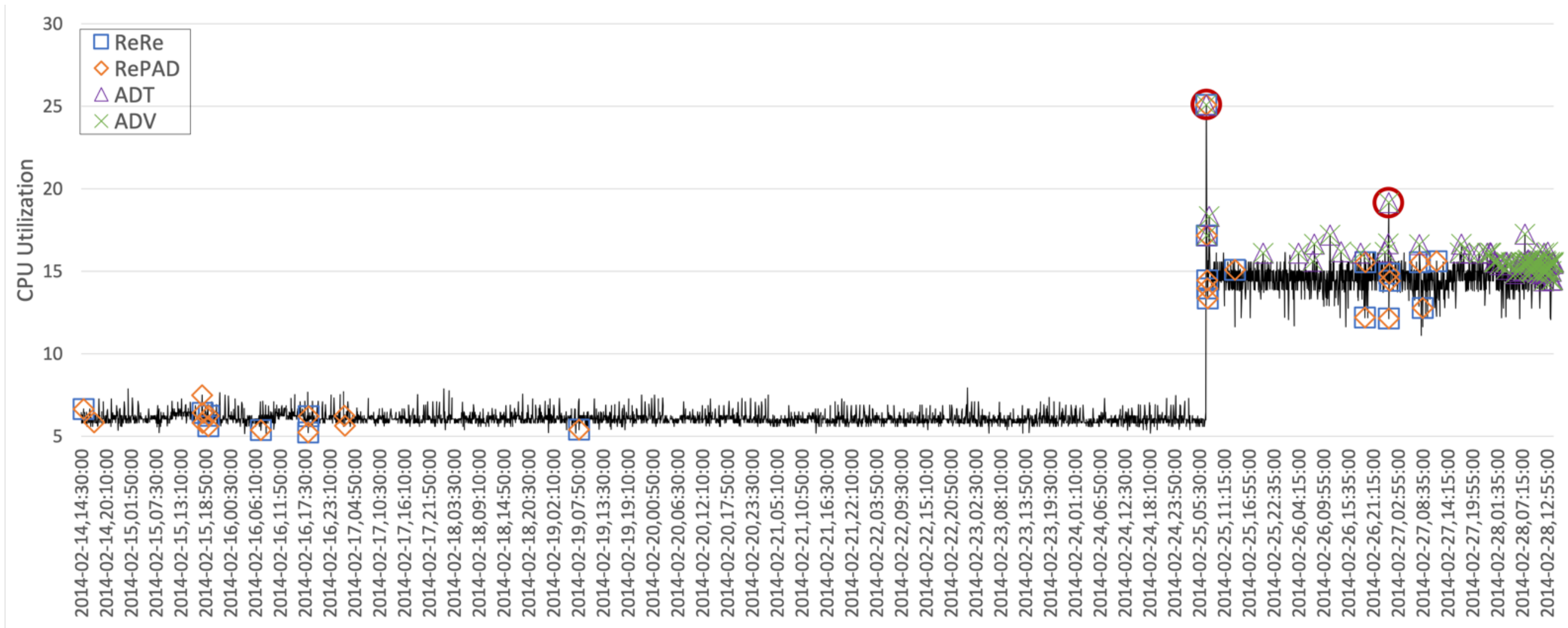

Fig. 6. The detection results of the four approaches on the CPU-c53 dataset. Note that this dataset has two anomalies labeled by human experts, marked as red circles.

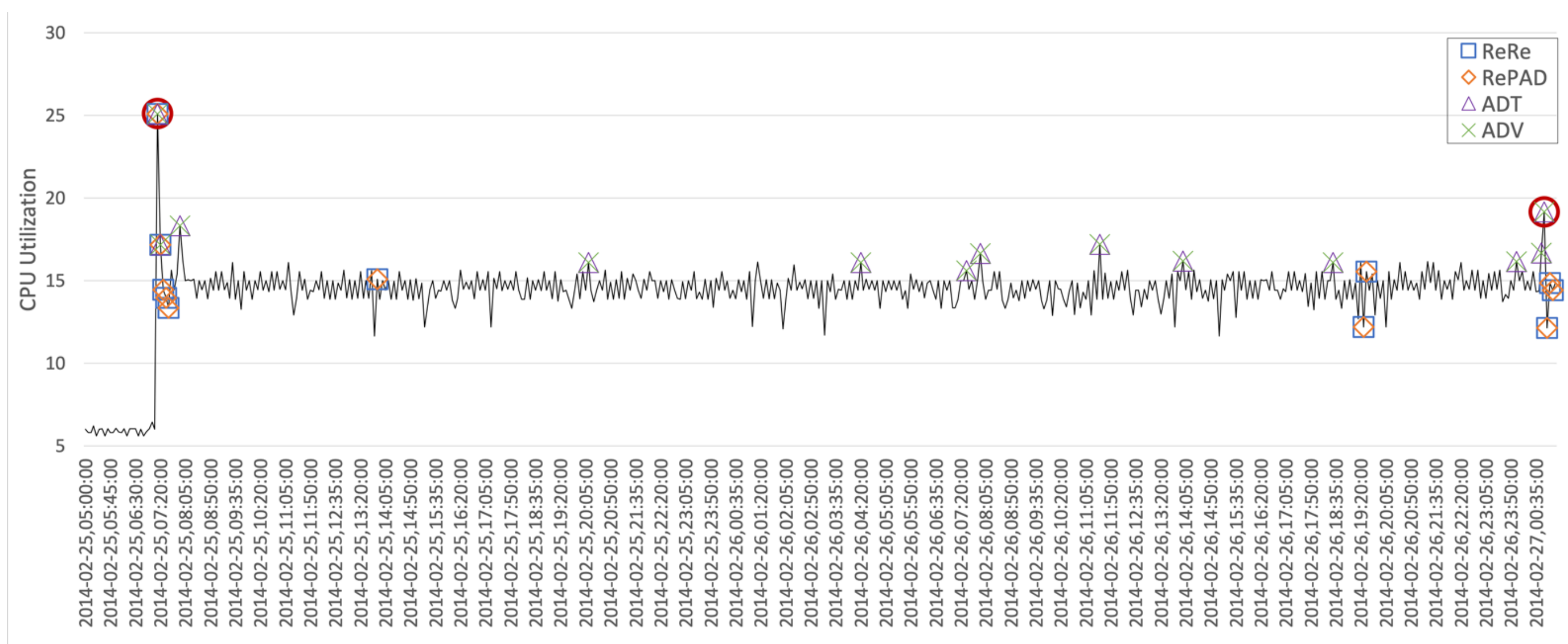

Fig. 7. A close-up of the detection results for the first and second anomalies on the CPU-c53 dataset.

TABLE VII. THE LSTM RETRAINING PERFORMANCE OF RERE AND REPAD ON THE CPU-C53 DATASET.

| Approach | # of data points that requires LSTM retraining | LSTM retraining ratio |
|---|---|---|
| ReRe | 111 | 2.76% (=111/4028) |
| RePAD | 59 | 1.46% (=59/4028) |

TABLE VIII. THE TIME CONSUMPTION OF RERE AND REPAD ON THE CPU-C53 DATASET

| Approach | Average Detection Time (sec) | Standard Deviation (sec) |
|---|---|---|
| ReRe | 0.018 | 0.032 |
| RePAD | 0.015 | 0.024 |

### C. *Experiment 3*

In the last experiment, we chose the real-world dataset real47 from the Yahoo benchmark datasets [15] to evaluate the four approaches since the anomalous data points in this dataset are considered particularly challenging to detect [11]. As illustrated in Fig. 8, this dataset contains 10 anomalous data points marked as red circles. Although these anomalous data points have normal values (except the last one), their shapes and patterns are unusual, which is the main reason why they are considered as anomalies by human experts.

As mentioned earlier, ADT is unable to execute on this dataset since ADT demands at least twice the amount of data points compared to what is present in real47. Hence, we are unable to measure the performance of ADT in this experiment. In addition, due to the short length of the real47 dataset, setting parameter *period* to be 1440 does not work for ADV. Hence, we evaluate the performance of ADV under three different settings (i.e., *period*=300, *period*=500, and *period*=700) for fair comparison.

Fig. 9 depicts a close-up of the detection results for the 10 anomalies in the real47 dataset. When ReRe and RePAD were separately employed, they made less false positives than ADV, implying that both ReRe and RePAD are able to learn the data distribution of the dataset and promptly adapt to pattern changes. According to TABLE IX, ReRe and RePAD provide the same superior detection performance in all the metrics when $K$=3 (which is a suggested measure according to [14] for any hourly-interval dataset). The result confirms that the short-term Look-Back and Predict-Forward strategy makes ReRe and RePAD able to detect these tough anomalies.

On the contrary, when ADV with *period*=300 was tested, it made neither true positives nor false positives since it considered all data points as normal. For this reason, the corresponding precision and recall are all zero, which makes it impossible to calculate the F-score based on Equation 1.

When ADV with period=500 was employed, the situation did not improve since ADV made several false positives for the last anomaly without being able to detect the other anomalies. Setting *period* as 700 for ADV seems a better choice since the corresponding precision, recall, and F-score increase. Nevertheless, ADV is still unable to outperform ReRe and RePAD.

TABLE IX. THE DETECTION PERFORMANCE OF RERE, REPAD, AND ADV ON THE REAL47 DATASET.

| Approach | Precision | | Recall | | F-score | |
|---|---|---|---|---|---|---|
| | $K$=0 | $K$=3 | $K$=0 | K=3 | $K$=0 | $K$=3 |
| ReRe | 0.125 | 0.7 | 0.1 | 1 | 0.111 | 0.824 |
| RePAD | 0.125 | 0.7 | 0.1 | 1 | 0.111 | 0.824 |
| ADV (*period* =300) | 0 | 0 | 0 | 0 | n/a | n/a |
| ADV (*period* =500) | 0 | 0 | 0 | 0 | n/a | n/a |
| ADV (*period* =700) | 0 | 0.308 | 0 | 0.036 | n/a | 0.064 |

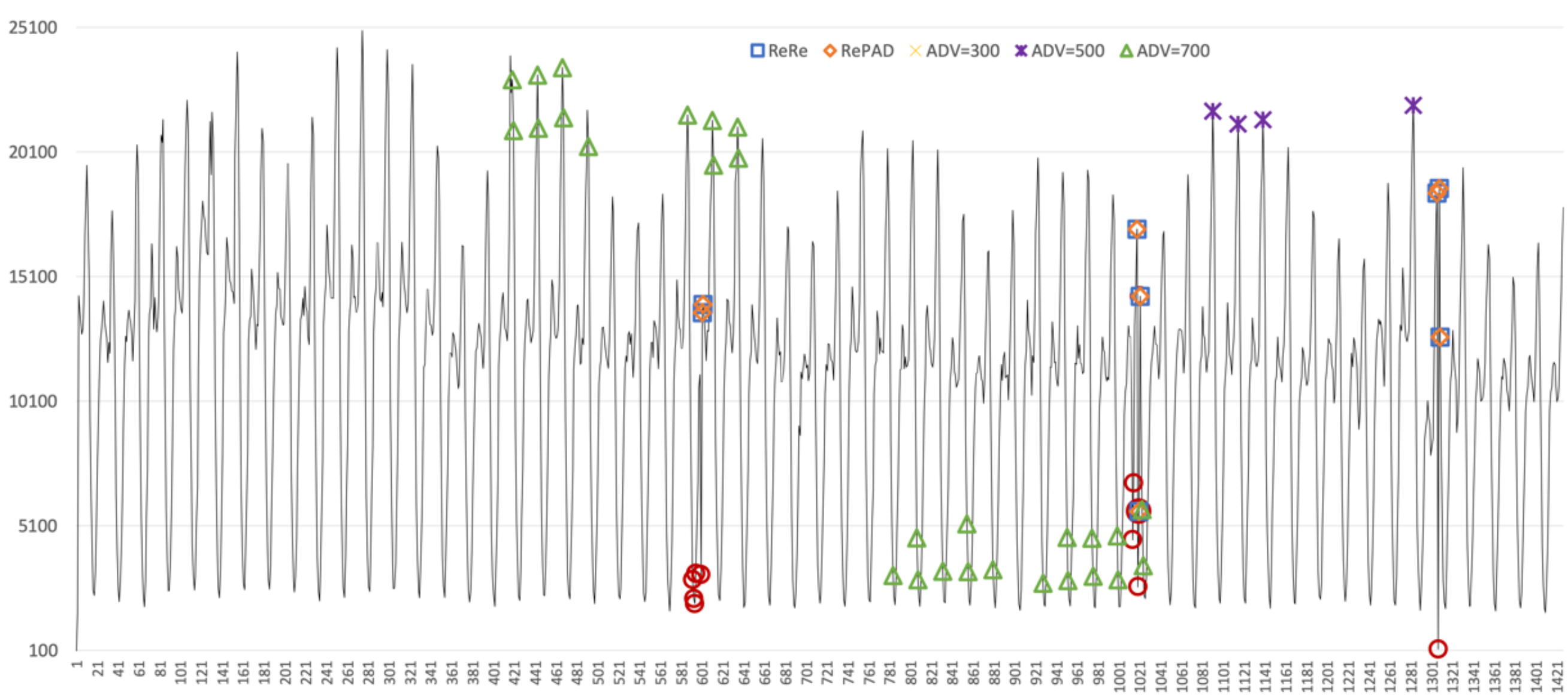

Fig. 8. The detection results of ReRe, RePAD, and ADV on the real47 dataset. Note that this dataset contains ten anomalies labeled by human experts, and they are marked as red circles.

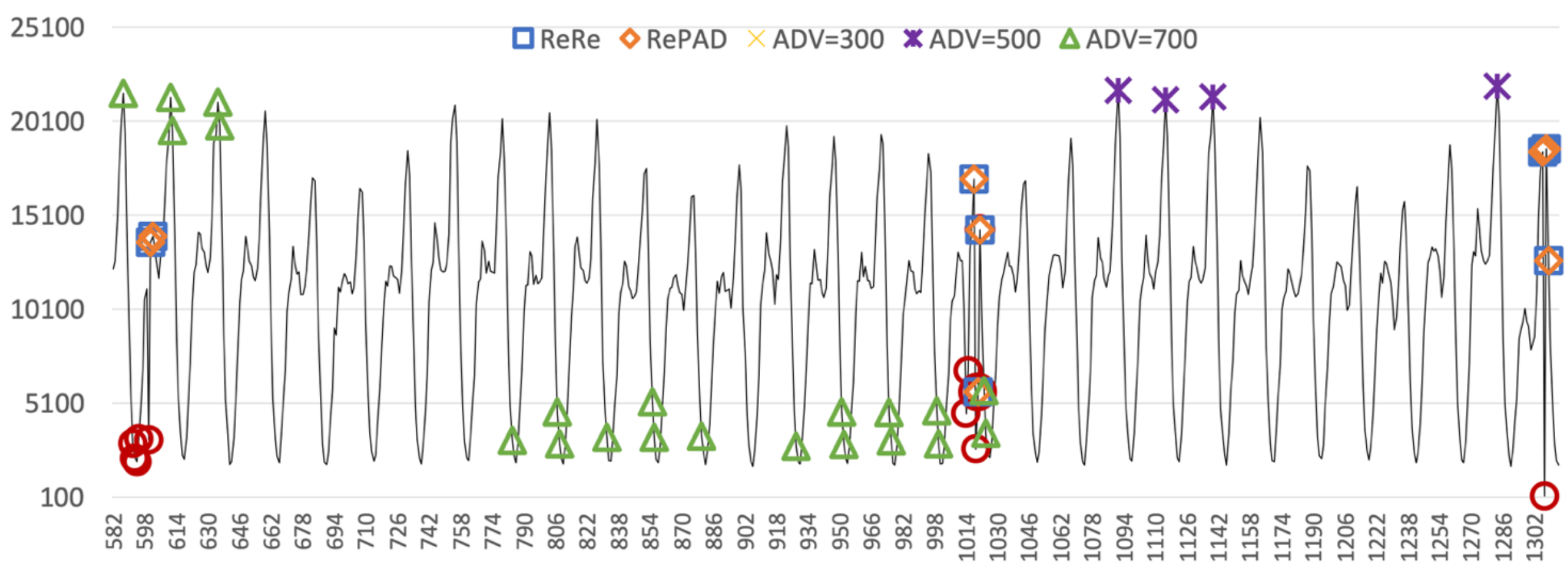

Fig. 9. A close-up of the detection results for ten anomalies on the real47 dataset.

TABLE X. THE LSTM RETRAINING PERFORMANCE OF RERE AND REPAD ON THE REAL47 DATASET.

| Approach | # of data points that requires LSTM retraining | LSTM retraining ratio |
|---|---|---|
| ReRe | 31 | 2.18% (=31/1422) |
| RePAD | 31 | 2.18% (=31/1422) |

TABLE XI. THE TIME CONSUMPTION OF RERE AND REPAD ON THE REAL47 DATASET.

| Approach | Average Detection Time (sec) | Standard Deviation (sec) |
|---|---|---|
| ReRe | 0.016 | 0.029 |
| RePAD | 0.015 | 0.028 |

TABLE X and XI summarize the time performance of ReRe and RePAD on detecting anomalies in the real47 dataset. Both ReRe and RePAD require to retrain their LSTM models at 31 data points, and the corresponding LSTM retraining ratio is low, only 2.8% (=31/1422), demonstrating the cost effectiveness of both ReRe and RePAD on the dataset. Due to the same reason, the average detection time required by these two approaches are also similar to each other, with similar standard deviations. The results show that even though ReRe employs one more LSTM with one more detection threshold, it is still very lightweight and able to conduct anomaly detection in real-time.

## V. CONCLUSION AND FUTURE WORK

In this paper, we have introduced ReRe for detecting anomalies in time series in a real-time manner. ReRe is able to work on any time series without needing to know the corresponding data distribution/patterns or data labels. In fact, ReRe requires no training data since it does not need to go through an off-line training process. This ready-to-go feature makes ReRe a practical solution in many real-world scenarios since it significantly reduces human effort.

After a very short probation period, ReRe starts its detection function without requiring a person to manually set detection thresholds. ReRe dynamically adjusts its two long-term detection thresholds over time and retrains its two LSTM models when necessary. These features enable ReRe to adapt to pattern changes in the target time series and detect anomalies in a time-efficient and real-time manner. Experiments based on real-world time series data demonstrate that ReRe provides satisfactory detection performance as compared with the other three state-of-the-art approaches. In addition, the lightweightness of ReRe makes it a cost-effective solution to be deployed on commodity machines.

As future work, we plan to further improve the detection performance of ReRe, especially in terms of false positives, by investigating hybrid solutions in a lightweight manner. In addition, we would like to extend ReRe for large-scale time series from the $eX^3$ HPC cluster [17] by referring to [18][19] and designing it in a parallel and distributed way.